\documentclass[11pt,a4paper]{article}
\usepackage[hyperref]{acl2019}
\usepackage{latexsym}
\usepackage{url}
\usepackage{times}
\usepackage{soul}
\usepackage{url}
\usepackage{graphicx}
\usepackage{amsmath}
\usepackage{booktabs}
\usepackage{times}
\usepackage{epsfig}
\usepackage{graphicx}
\usepackage{amsmath}
\usepackage{amssymb}
\usepackage{bm}
\usepackage{cuted,booktabs}
\usepackage{todonotes}

\aclfinalcopy 

\title{Faithful Multimodal Explanation for Visual Question Answering}

\author{Jialin Wu\\
  Department of Computer Science \\
  University of Texas at Austin \\
  \texttt{jialinwu@cs.utexas.edu} \\ \And
  Raymond J. Mooney \\
    Department of Computer Science \\
  University of Texas at Austin \\
  \texttt{mooney@cs.utexas.edu}}

\date{}

\begin{document}
\maketitle
\begin{abstract}
 AI systems' ability to explain their reasoning is critical to their utility and trustworthiness.  Deep neural networks have enabled significant progress on many challenging problems such as visual question answering (VQA). However, most of them are opaque black boxes with limited explanatory capability. This paper presents a novel approach to developing a high-performing VQA system that can elucidate its answers with integrated textual and visual explanations that faithfully reflect important aspects of its underlying reasoning process while capturing the style of comprehensible human explanations. Extensive experimental evaluation demonstrates the advantages of this approach compared to competing methods using both automated metrics and human evaluation.
\end{abstract}
\section{Introduction}
Deep neural networks have made significant progress on visual question answering (VQA), the challenging AI problem of answering natural-language questions about an image \cite{antol2015vqa}.  
However successful systems \cite{fukui2016multimodal,anderson2017bottom,yang2016stacked,wu2018joint,jiang2018pythia} based on deep neural networks are difficult to comprehend because of many layers of abstraction and a large number of parameters. This makes it hard to develop user trust.
Partly due to the opacity of current deep models, there has been a recent resurgence of interest in {\it explainable AI}, systems that can explain their reasoning to human users.  In particular, there has been some recent development of explainable VQA systems \cite{selvaraju2017grad,park2018multimodal,hendricks2016generating,hendricks2018grounding}.

\begin{figure}
    \centering
    \includegraphics[width=\linewidth, trim={4.5cm 6cm 11cm 1.5cm}, clip]{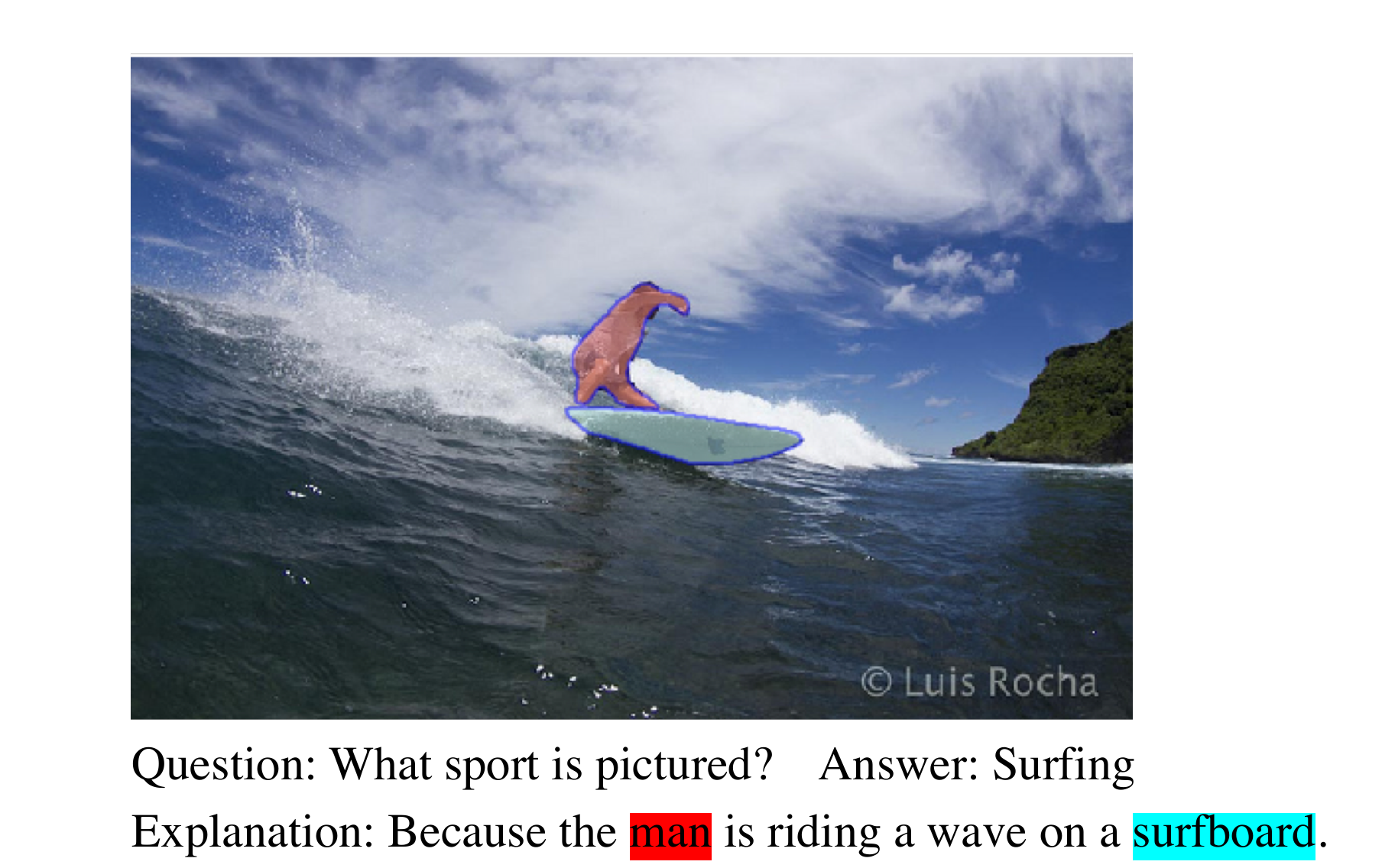}
    \caption{Example of our multimodal explanation. It highlights relevant image regions together with a textual explanation with corresponding words in the same color.}
    \label{fig:sample-explanation}
\end{figure}
One approach to explainable VQA is to generate {\it visual explanations}, which highlight image regions that most contributed to the system's answer, as determined by attention mechanisms \cite{lu2016hierarchical} or gradient analysis \cite{selvaraju2017grad}. However, such simple visualizations do not explain {\it how} these regions support the answer. An alternate approach is to generate a {\it textual explanation}, a natural-language sentence that provides reasons for the answer. Some recent work has generated textual explanations for VQA by training a recurrent neural network (RNN) to directly mimic examples of human explanations \cite{hendricks2016generating,park2018multimodal}. A {\it multimodal} approach that integrates {\it both} a visual and textual explanation provides the advantages of both.  Words and phrases in the text can point to relevant regions in the image. An illustrative explanation generated by our system is shown in Figure. \ref{fig:sample-explanation}.

Recent research on such multimodal VQA explanation is presented in \cite{park2018multimodal} that employs a form of ``post hoc justification'' that does not truly follow and reflect the system's actual processing. We believe that explanations should more faithfully reflect the actual processing of the underlying system in order to provide users with a deeper understanding of the system, increasing trust for the right reasons, rather than trying to simply convince them of the system's reliability \cite{bilgic:iui05-wkshp}. In order to be faithful, the textual explanation generator should focus on the set of objects that contribute to the predicted answers, and receive proper supervision from only the gold standard explanations that are consistent with the actual VQA reasoning process. Towards this end, our explanation module directly uses the VQA-attended features and is trained to only generate human explanations that can be traced back to the relevant object set using a gradient-based method called GradCAM \cite{selvaraju2017grad}. To enforce local faithfulness, we also align the gradient-based visual explanations generated by the textual explanation module and the VQA module during training.

In addition, our explanations provide direct links between terms in the textual explanation and segmented items in the image, as shown in Figure \ref{fig:sample-explanation}.
The result is a nice synthesis of a faithful explanation that highlights concepts actually used to compute the answer and a comprehensible, human-like, linguistic explanation. Below we describe the details of our approach and present extensive experimental results on the VQA-X \cite{park2018multimodal} dataset that demonstrates the advantages of our approach compared to prior work using this data \cite{park2018multimodal} in terms of both automated metrics and human evaluation. Further, in order to evaluate the faithfulness, we design two metrics: (1) We first measure the degree of similarity between the highlighted image segments in our textual explanations and the influential segments determined by the LIME explainer \cite{lime}; (2) we also measure the consistency between the gradient-based visual explanation \cite{selvaraju2017grad} of the predicted answer and the generated textual explanation. 

\begin{figure*}
    \centering
    \includegraphics[width=1.1\linewidth, trim={2cm 9.5cm 7.5cm 1.5cm}, clip]{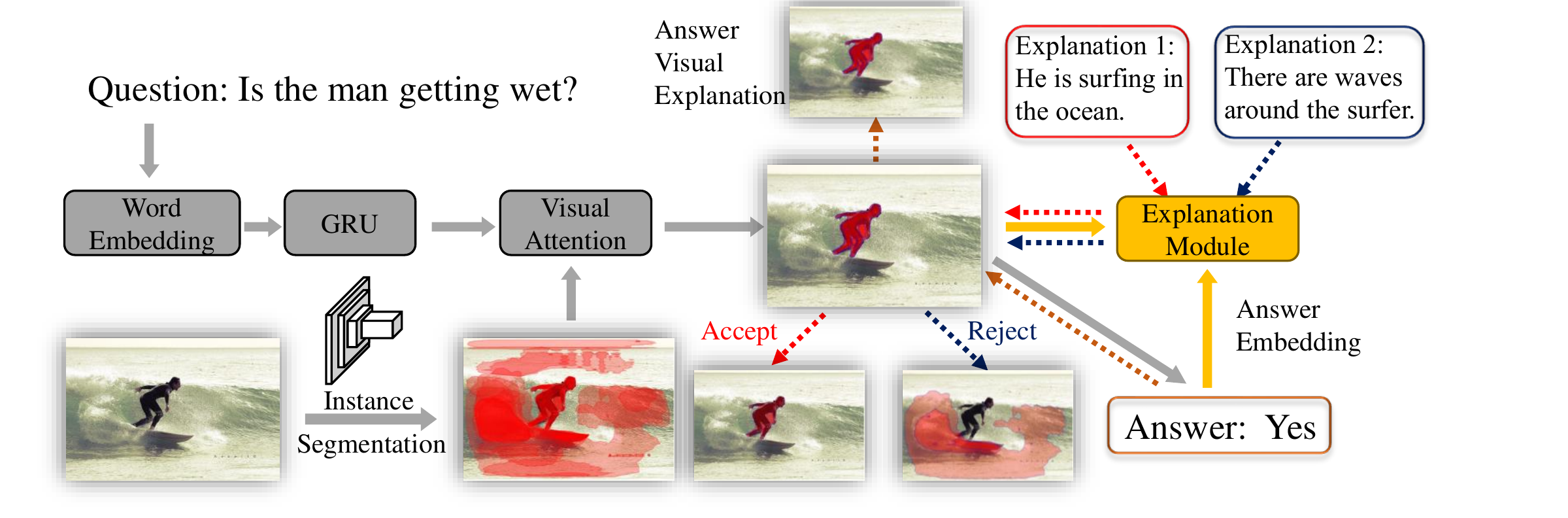}
    \caption{Model overview: We first segment the image and then predict the answer for the visual question with a pretrained VQA module. Then, we learn to embed the question, answer, and the VQA-attended features to generate textual explanations. During training, we only use the faithful human explanation whose gradient-based visual explanation is consistent with that of the predicted answer. In the example, our explanation module is only trained to generate ``Explanation 1'' and further enforces the consistency between this explanation and the predicted answer.  ``Explanation 2'' is filtered out since its visual explanation is mainly focused on the waves and is not consistent with VQA module's focus on the surfer. Dashed arrows denote gradients, gray and yellow arrows denote fixed and trainable parameters, respectively. The three smaller images denote the visual explanations for the predicted answer and the two textual explanations.}
    \label{fig:overview}
\end{figure*}

\section{Related Work}
In this section, we review related work including visual and textual explanation generation and VQA.
\subsection{VQA}
Answering visual questions \cite{antol2015vqa} has been widely investigated in both the NLP and computer vision communities. Most VQA models \cite{fukui2016multimodal,lu2016hierarchical} embed images using a CNN and questions using an RNN  and then use these embeddings to train an answer classifier to predict answers from a pre-extracted set. Attention mechanisms are frequently applied to recognize important visual features and filter out irrelevant parts. 
A recent advance is the use of the Bottom-Up-Top-Down (Up-Down) attention mechanism \cite{anderson2017bottom} that attends over high-level objects instead of convolutional features to avoid emphasis on irrelevant portions of the image. We adopt this mechanism, but replace object detection \cite{girshick2015fast} with segmentation \cite{hu2018learning} to obtain more precise object boundaries.

\subsection{Visual Explanation}
A number of approaches have been proposed to visually explain decisions made by vision systems by
highlighting relevant image regions. Grad-CAM \cite{selvaraju2017grad} analyzes the gradient space to find visual regions that most affect the decision. 
Attention mechanisms in VQA models can also be directly used to determine highly-attended regions and generate visual explanations. Unlike conventional visual explanations, ours highlight segmented objects that are linked to words in an accompanying textual explanation, thereby focusing on more precise regions and filtering out noisy attention weights.

\subsection{Textual and Multimodal Explanation}
Visual explanations highlight key image regions behind the decision, however, they do not explain the reasoning process and crucial relationships between the highlighted regions. 
Therefore, there has been some work on generating textual explanations for decisions made by visual classifiers \cite{hendricks2016generating}. As mentioned in the introduction, there has also been some work on multimodal explanations that link textual and visual explanations \cite{park2018multimodal}.  A recent extension of this work \cite{hendricks2018grounding} first generates multiple textual explanations and then filters out those that could not be grounded in the image. We argue that a good explanation should focus on referencing visual objects that actually influenced the system's decision, therefore generating more faithful explanations.

\section{Approach}
Our goal is to generate more faithful multimodal explanations that specifically include the segmented objects in the image that are the focus of the VQA module. Figure \ref{fig:overview} illustrates our model's pipeline in the training phase, consisting of the VQA module (Section \ref{sec:vqa}), and textual explanation module (Section \ref{sec:explanation}). We first segment the objects in the image and predict the answer using the VQA module, which has an attention mechanism over those objects. Next, the explanation module is trained to generate textual explanations conditioned on the question, answer, and VQA-attended features. To faithfully train the explanation module, we filter out human textual explanations whose gradient-based visual explanation is not consistent with that of the predicted answer.
For example, in Figure \ref{fig:overview} ``Explanation 1'' is accepted as the textual explanation since it is mainly focused on the surfer and ``Explanation 2'' is rejected. For the consistent textual explanations, we also train the explanation module to align its visual explanation with the predicted answer's to enforce local faithfulness. 

\subsection{Notation}
We use $f$ to denote the fully-connected  $fc$ layers of the neural network, and these $fc$ layers do not share parameters.  We notate the sigmoid functions as $\sigma$. The subscript $i$ indexes the elements of the segmented object sets from images. Bold letters denote vectors, overlining $\overline{\cdot}$ denotes averaging, and [$\cdot$, $\cdot$] denotes concatenation.

\subsection{VQA Module} 
\label{sec:vqa}
We base our VQA module on Up-Down \cite{anderson2017bottom} with some modifications. 
First, we replace the two-branch gated \textit{tanh} answer classifier with single \textit{fc} layers with Leaky ReLU activation \cite{maas2013rectifier}. In order to ground the explanations in more precise visual regions, we use instance segmentation \cite{hu2018learning} to segment objects in over 3,000 categories. Specifically, we extract at most the top $V<80$ objects in terms of segmentation scores and concatenate each object's \textit{fc6} representation in the bounding box classification branch and \textit{mask\_fcn[1-4]} features in the mask generation branch to form a 2048-d vector. This results in an image feature set $\textbf{V}$ containing $V$ 2048-d vectors $\textbf{v}_i$ for each image. We encode each question as the last hidden state \textbf{q} of a gated recurrent unit (GRU) with $512$ hidden units. We learn visual attention over all the segments $\bm{\alpha}^{vqa}\in \mathbb{R}^{V}$, and use the attended visual features  $\textbf{v}^q_i$ together with the question embedding to produce a joint representation $\textbf{h}$. Then the model predicts the logits $\textbf{s}^{vqa}$ for each answer candidate using a 2-layer $fc$ networks, which is passed through a sigmoid function to compute the final probabilities. For the detailed network architecture, please refer to \cite{anderson2017bottom}. The parameters in the VQA module are fixed during the training of the explanation module.

\subsection{Question and Answer Embedding for Explanation Generation} 
As suggested in \cite{park2018multimodal}, we also encode questions and answers as input features to the explanation module. In particular, we regard the normalized answer prediction output as a multinomial distribution, and sample one answer from this distribution at each time step. We re-embed it as a one-hot vector $\textbf{a}_s=\text{one-hot} (\text{multinomial}(s))$.

\begin{align}
    \textbf{u}_i &= \textbf{v}^q_i \odot f(\textbf{a}_s) \odot f(\textbf{q})\label{eq:qa_embed}
\end{align}
Next, we element-wise multiply the embedding of $\textbf{q}$ and $\textbf{a}_s$ with  $\textbf{v}^q_i$ to compute the joint representation $\textbf{u}_i$. Note that $\textbf{u}$ faithfully represents the focus of the VQA process, in that it is directly derived from the VQA-attended features.

\subsection{Explanation Generation} 
\label{sec:explanation}
In this section, we describe the Explanation Module depicted by the yellow box in Figure.\ \ref{fig:overview}. The explanation module has a two-layer-LSTM architecture whose first layer produces an attention over the $\textbf{u}_i$, and whose second layer learns a representation for predicting the next word using the first layer's features. \\
\begin{figure}[h]
    \centering
    \includegraphics[width=\linewidth, trim={0cm 11cm 24cm 0.cm}, clip]{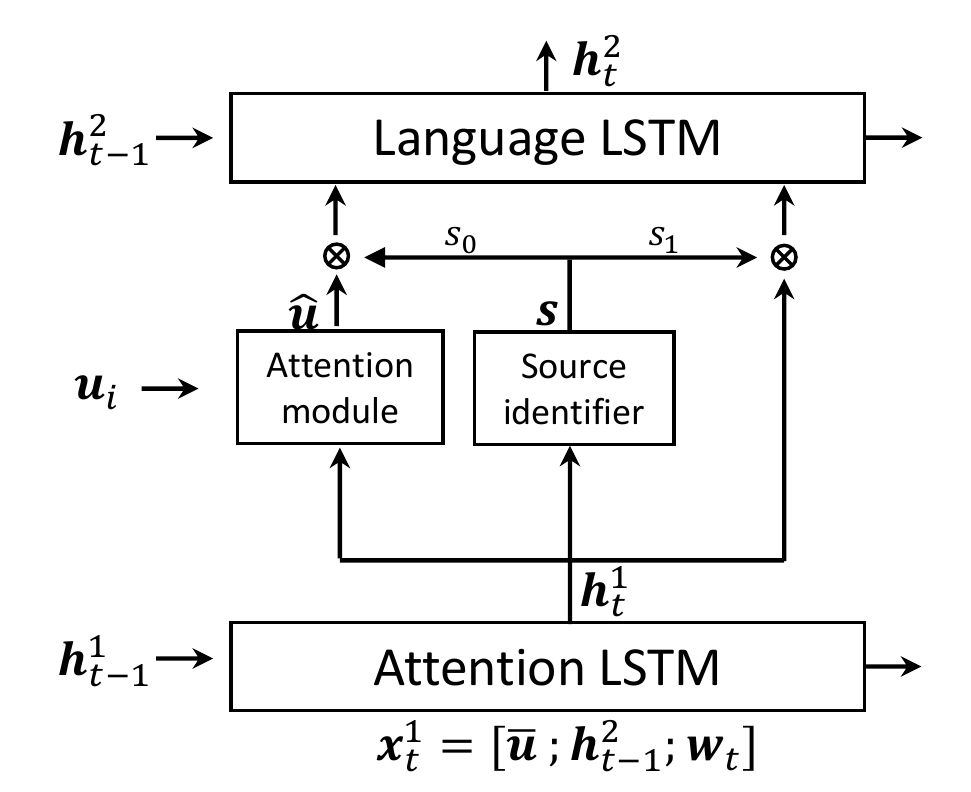}
    \caption{Overview of the explanation module.}
    \label{fig:explanation_module}
\end{figure}

In particular, the first visual attention LSTM takes the concatenation $\textbf{x}_t^1$ of the second language LSTM's previous output $\textbf{h}^2_{t-1}$, the average pooling of $\textbf{u}_i$, and the previous words' embedding as input and produces the hidden presentation $\textbf{h}^1_t$. Then, an attention mechanism re-weights the image feature $\textbf{u}_i$ using the generated $\textbf{h}^1_t$ as input shown in Eq. \ref{eq:att_lstm}. For the detailed structure, please refer to \cite{anderson2017bottom}.

\begin{align}
\label{eq:att_lstm}
a_{i,t} &= f(tanh (f(\textbf{u}_i) + f(\textbf{h}^1_t)))\\
\bm{\alpha}_t &= softmax(\bm{\alpha}_t)
\end{align}

For the purpose of faithfully grounding the generated explanation in the image, we argue that the generator should be able to determine if the next word should be based on image content attended to by the VQA system or on learned linguistic content. To achieve this, we introduce a ``source identifier'' to balance the total amount of attention paid to the visual features $\textbf{u}_i$ and the recurrent hidden representation $\textbf{h}^1_t$ at each time step. In particular, given the output $\textbf{h}^1_t$ from the attention LSTM and the average pooling $\overline{\textbf{u}}_i$ over $\textbf{u}_i$, we train a $fc$ layer to produce a 2-d output $\textbf{s} = \sigma(f([\textbf{h}^1_t, \ \overline{\textbf{u}}_i])) = (s_0, s_1)$ that identifies which source the current generated word should be based on ($i.e.$ $s_0$ for the output of the attention LSTM\footnote{We tried to directly use the source weights $s_0$ in the language LSTM's hidden representation $\textbf{h}^2_{t-1}$ and found that using $\textbf{h}^1_{t}$ works better. The reason is that directly constraining $\textbf{h}^2_{t-1}$ makes the language LSTM forget  the previously encoded content and prevents it from learning long term dependencies.} and $s_1$ for the attended image features). 
\begin{align}
\label{eq:source_identifier}
    \textbf{s} = \sigma(f([\textbf{h}^1_t, \ \overline{\textbf{u}}_i]))
\end{align}

We use the following approach to obtain training labels $\hat{\textbf{s}}$ for the source identifier. For each visual features $\textbf{u}_i$, we assign label $1$ (indicating the use of attended visual information) when there exists a segmentation $\textbf{u}_i$ whose cosine similarity between its category name's GloVe representation and the current generated word's GloVe representation is above $0.6$.  Given the labeled data, we train the source identifier using cross entropy loss $\mathcal{L}_{s}$ as shown in Eq. \ref{eq:source_loss}:
\begin{align}
    \mathcal{L}_{s} = -(\sum^1_{j=0} \hat{s}_{j}\log s_{j} + (1-\hat{s}_{j})\log(1-s_{j}))
    \label{eq:source_loss}
\end{align}
where the $\hat{s}_0,\hat{s}_1$ are the aforementioned labels. 

Next, we concatenate the re-weighted $\textbf{h}^1_t$ and $\overline{\textbf{u}}_i$ with the output of the source identifier as the input $\textbf{x}^2_t = [\textbf{h}^1_ts_0, \ \overline{\textbf{u}}_i s_1]$ for the language LSTM. For more detail on the language LSTM structure, please refer to \cite{anderson2017bottom}. 

With the hidden states $\textbf{h}^2_t$ in the Language LSTM, the output word's probability is computed using Eq. \ref{eq:con_prob}:
\begin{align}
\label{eq:con_prob}
p(y_t | y_{1:t-1}) = softmax (f(\textbf{h}^2_t))
\end{align}
where $y_t$ denotes the $t$-th word in the explanation $\textbf{y}$ and $y_{1:t-1}$ denotes the first $t-1$ words.\\

\noindent\textbf{Faithful Explanation Supervision.} Directly collecting faithful textual explanations is infeasible because it would require an annotation process where workers provide explanations based on the attended VQA features. Instead, we design an online algorithm that automatically filters unfaithful explanations from the human ones in the VQA-X data \cite{park2018multimodal}  based on the idea that a proper explanation should focus on the same  set of objects as the VQA module and be locally faithful.  As recent research suggested that gradient-based methods more faithfully present the models' decision making process \cite{zhang2018top,wu2018dynamic,wu2018self,jain2019attention}, we define a faithfulness score $\mathcal{S}_f$ as the cosine similarity between the Grad-CAM \cite{selvaraju2017grad} visual explanation vectors of the textual explanation and the predicted answer as shown in Eq. \ref{eq:faith_score}:
\begin{align}
\mathcal{S}_f(\textbf{y}) = cos(g(s^{vqa}_{pred}, \textbf{v}^q), g(\log p(\textbf{y}), \textbf{v}^q)) \label{eq:faith_score}
\end{align}
where $g$ denotes the Grad-CAM operation and the result is a vector of length $|V|$ indicating the contribution of each segmented object. $s^{vqa}_{pred}$ is the logit for the predicted answer.

Then, we filter out the explanations in the training set whose faithfulness scores are less than $\xi \max(0.02 \ it, 1)$, where $\xi$ is a threshold and the $\max(0.02 \ it, 1)$ term is used to jump-start the randomly initialized explanation module. For example, during training, we only accept ``Explanation 1'' in Figure \ref{fig:overview} because the visual explanations of the predicted answer and the textual explanation are consistent and reject ``Explanation 2''.

Since the VQA-X dataset only has explanations for the correct answers, we also discard the explanations when the predicted answers are wrong. 
With the remaining human explanations, we minimize the cross-entropy loss $\mathcal{L}_{XE}$ in Eq. \ref{eq:xe}:
\begin{align}
\mathcal{L}_{XE} &= − \sum_{t=1}^{T}\log(p(y_t |y_{1:t-1})) \label{eq:xe}
\end{align}
To enforce local faithfulness, we further align these two gradient vectors using cosine distance $\mathcal{L}_{f}=1-\mathcal{S}_f$.

\begin{figure}[h]
    \centering
    \includegraphics[width=\linewidth, trim={1.5cm 13.4cm 11cm 1.5cm}, clip]{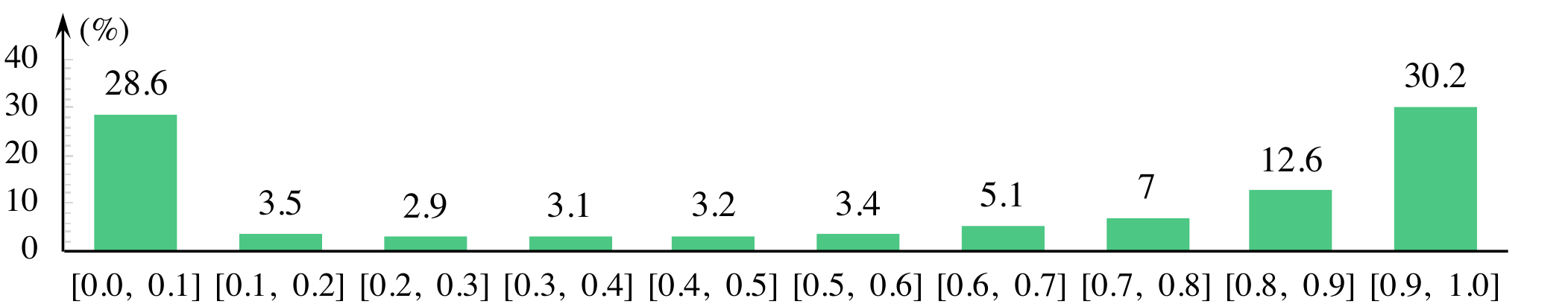}
    \caption{The distribution of explanations' faithfulness scores in the last epoch during training.}
    \label{fig:train_distribution}
\end{figure}

In Figure \ref{fig:train_distribution}, we report the distribution of the explanations' faithfulness scores $\mathcal{S}_f$ in the last epoch during training ($\xi$ is set to $0.3$). We observe that about 30\% of the human explanations in the training set cannot be traced back to similar image segments that highly contribute to the predicted answer using our trained explanation module. These textual explanations cannot be seen as faithful either because the explanations themselves are not faithful or because the explanation module fails to develop the correct mappings between the textual explanations and the VQA-attended features. There are only a small fraction of the explanations whose faithfulness scores are in the interval of [0.1, \ 0.6] indicating that there is a clear boundary between whether or not an explanation is deemed faithful according to our metric.

\begin{table*}[!t]
\centering
\begin{tabular}{l|ccc|c|ccccc|c}
\toprule
        & && &  & \multicolumn{5}{c|}{Textual}               & Visual  \\\hline
         &$\mathcal{L}_{s}$&$\mathcal{F}$&$\mathcal{L}_{f}$&\# Expl. & B-4 & M & R-L & C & S  & EMD                            \\ \hline\hline
PJ-X \cite{park2018multimodal}& & & &  29K  & 19.5   & 18.2   & 43.7  & 71.3  & 15.1  &  2.64   \\
Ours (Justification) & & & & 29K   & 23.5   & 19.0   & 46.2  & 81.2  & 17.2   & 2.46    \\
Ours (Justification) &\checkmark & & &  29K  & 24.4   & 19.5   & 47.4  & 88.8  & 17.9  & 2.41\\ \hline
Ours (Justification) &\checkmark & & & 15K   & 24.1   & 18.6   & 46.2  & 83.4  & 16.2   & 2.59 \\
Ours (Explanation) &\checkmark &\checkmark & & 15K  & 24.7  & 19.2   & 47.0  & 85.1  & 16.6   & 2.56     \\
Ours (Explanation)  &\checkmark&\checkmark&\checkmark & 15K &\textbf{ 25.1  } & \textbf{19.7}   & \textbf{48.2}  & \textbf{86.7}  & \textbf{17.2}   & \textbf{2.52}     \\\bottomrule               
\end{tabular}
\caption{Explanation evaluation results, the top half shows results using the entire train set and the bottom half shows results using about 15K explanations. $\mathcal{F}$ denotes whether to filter out the unfaithful training explanations. With $\mathcal{F}$, the 15K explanations are the remaining explanation and without $\mathcal{F}$, the 15K explanations are randomly sampled from train set. $\mathcal{L}_s$, $\mathcal{L}_f$ denote the losses of the source identifier and the faithful gradient alignment, respectively. B-4, M, R-L, C and S are short hand for BLEU-4, METEOR, ROUGE-L, CIDEr and SPICE, respectively.}
\label{tab:results}
\end{table*}

\subsection{Training}
We pre-train the VQA module on the entire VQA v2 training set for $15$ epochs using the Adam optimizer \cite{kingma2014adam} with a learning rate of 0.001. After that, the parameters in the VQA module are frozen. Our VQA module is capable of achieving 82.9\% and 80.3\% in the VQA-X train and test split respectively. and 63.5\% in the VQA v2 validation set which is comparable to the baseline Up-Down model (63.2\%) \cite{anderson2017bottom}. Note that VQA performance is not the focus of this work, and our
experimental evaluation focuses on the generated explanations. Finally, we train the explanation module using the human explanations in the VQA-X dataset \cite{park2018multimodal} filtered for faithfulness. VQA-X contains 29,459 question answer pairs and each pair
is  associated  with  a  human  explanation. We train to minimize the joint loss $\mathcal{L}$ (Eq. \ref{eq:joint_minimization}), and $\xi$ is empirically set to $0.3$. We ran the Adam optimizer for 25 epochs with a batch size of 128. The learning rate for training the explanation module is initialized to 5e-4 and decays by a factor of 0.8 every three epochs.
\begin{align}
\label{eq:joint_minimization}
\mathcal{L} = \mathcal{L}_{XE}+ \mathcal{L}_{s} +\mathcal{L}_{f}
\end{align}
\subsection{Multimodal Explanation Generation}
\label{sec:multimodal}
As a last step, we link words in the generated textual explanation to image segments in order
to generate the final multimodal explanation. To determine which words to link, we extract all common nouns whose source identifier weight $s_1$ in Eq.\ \ref{eq:source_identifier} exceeds 0.5.  We then link them to the segmented object with the highest attention weight $\bm{\alpha}_t$  in Eq. \ref{eq:att_lstm} when that corresponding output word $y_t$ was generated, but only if this weight is greater than 0.2.\footnote{Due to duplicated segments, we use a lower threshold.}

\section{Experimental Evaluation}
This section experimentally evaluates both the textual and visual aspects of our multimodal explanations, including comparisons to competing methods and ablations that study the impact of the various components of our overall system. Finally, we present metrics and evaluation for the faithfulness of our explanations. 

\subsection{Textual Explanation Evaluation}
Similar to \cite{park2018multimodal}, we first evaluate our textual explanations using automated metrics by comparing them to the gold-standard human explanations in the VQA-X test data using standard sentence-comparison metrics: BLEU-4 \cite{Papineni:2002:BMA:1073083.1073135}, METEOR \cite{banerjee2005meteor}, ROUGE-L \cite{lin2004rouge}, CIDEr \cite{vedantam2015cider} and SPICE \cite{spice2016}. Table \ref{tab:results} reports our performance, including ablations. In particular, ``Justification'' denotes training on the entire or randomly sampled VQA-X dataset and ``Explanation'' denotes training only on the remaining faithful explanations. We outperform the current state-of-the-art PJ-X model \cite{park2018multimodal} on all automated metrics by a clear margin with only about half the explanation training data. This indicates that constructing explanations that  faithfully reflect the VQA process can actually generate explanations that match human explanations better than just training to directly match human explanations, possibly by avoiding over-fitting and focusing more on important aspects of the test images.\\

\subsection{Multimodal Explanation Evaluation}
In this section, we present the evaluations of our model on both visual and multimodal aspects.\\

\noindent\textbf{Automated Evaluation:} 
As in previous work \cite{selvaraju2017grad,park2018multimodal}, we first used Earth Mover Distance (EMD) \cite{pele2008}
to compare the image regions highlighted in our explanation to image regions highlighted by human judges.
In order to fairly compare to prior results, we resize all the images in the entire test split to 14$\times$14 and adjust the segmentation in the images accordingly using bi-linear interpolation. Next, we sum up the multiplication of attention values and source identifiers' values in Eq, \ref{eq:att_lstm} over time ($t$) and assign the accumulated attention weight to each corresponding segmentation region. We then normalize attention weights over the 14 $\times$ 14 resized images to sum to 1, and finally compute the EMD between the normalized attentions and the ground truth.    

As shown in the Visual results in Table \ref{tab:results}, our approach matches human attention maps more closely than PJ-X \cite{park2018multimodal}. We attribute this improvement to the following reasons. First, our approach uses detailed image segmentation which avoids focusing on background and is much more precise than bounding-box detection. Second, our visual explanation is focused by textual explanation where the segmented visual objects must be linked to specific words in the textual explanation. Therefore, the risk of attending to unnecessary objects in the images is significantly reduced. As a result, we filter out most of the noisy attention in a purely visual explanation like that in PJ-X.
\begin{figure}[!t]
    \centering
    \includegraphics[width=\linewidth, trim={0cm 11.5cm 11cm 1cm}, clip]{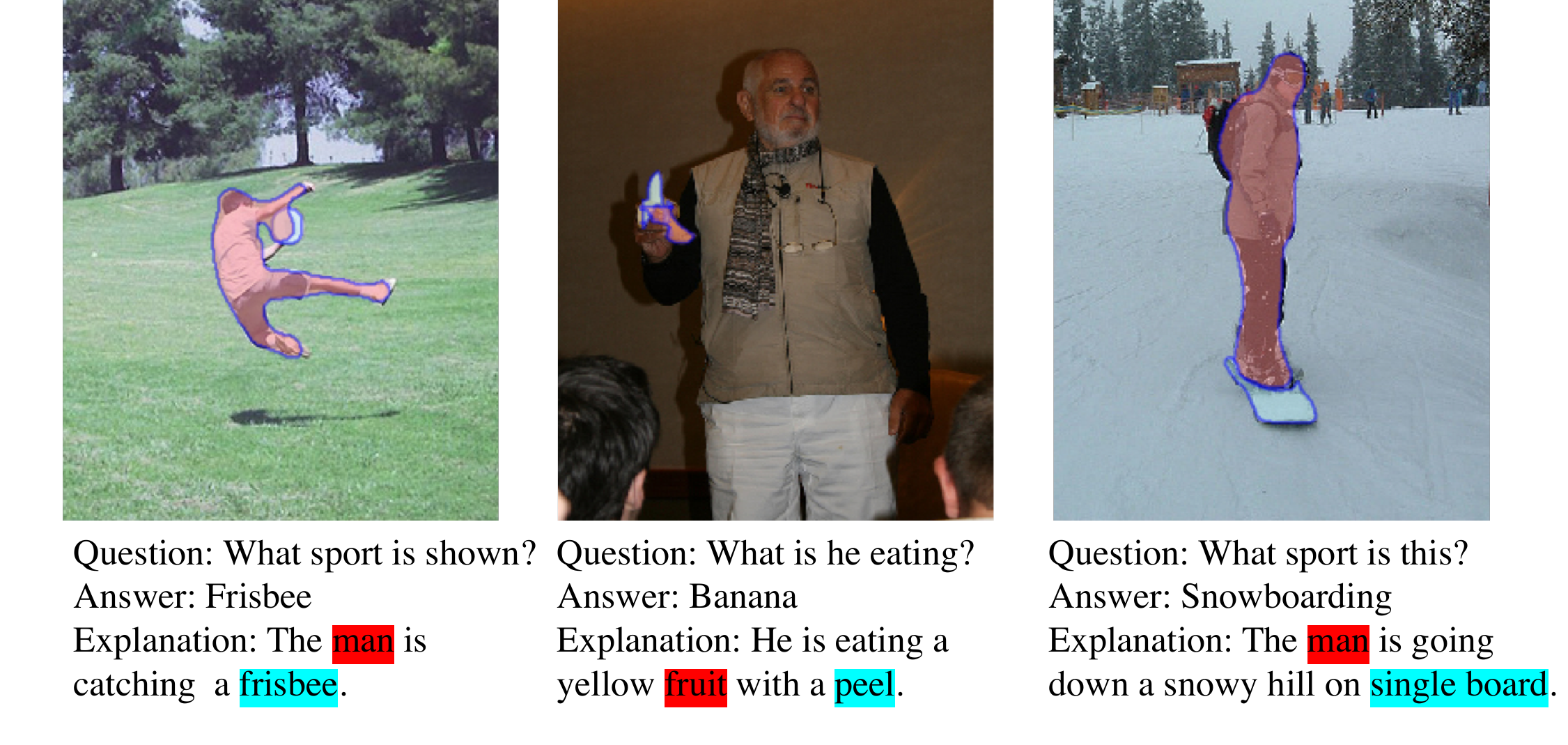}
    \caption{Sample positively-rated explanations. }
    \label{fig:examples_cheery}
\end{figure}

\noindent\textbf{Human Evaluation:} We also asked AMT workers to evaluate our final multimodal explanations that link words in the textual explanation directly to segments in the image.
Specifically, we randomly selected 1,000 correctly answered question and asked workers `` How well do the highlighted image regions support the answer to the question?''  and provided them a Likert-scale set of possible answers: ``Very supportive'', ``Supportive'', ``Neutral'', `Unsupportive'' and ``Completely unsupportive''. The second task was to evaluate the quality of the links between words and image regions in the explanations. We asked workers ``How well do the colored image segments highlight the appropriate regions for the corresponding colored words in the explanation?'' with the Like-scale choices: ``Very Well'', ``Well'', ``Neutral'', ``Not Well'', ``Poorly''.
We assign five questions in each AMT HIT with one ``validation'' item to control the HIT's qualities. 
\begin{figure}[h]
    \centering
    \includegraphics[width=\linewidth, trim={0cm 10cm 7cm 0cm}, clip]{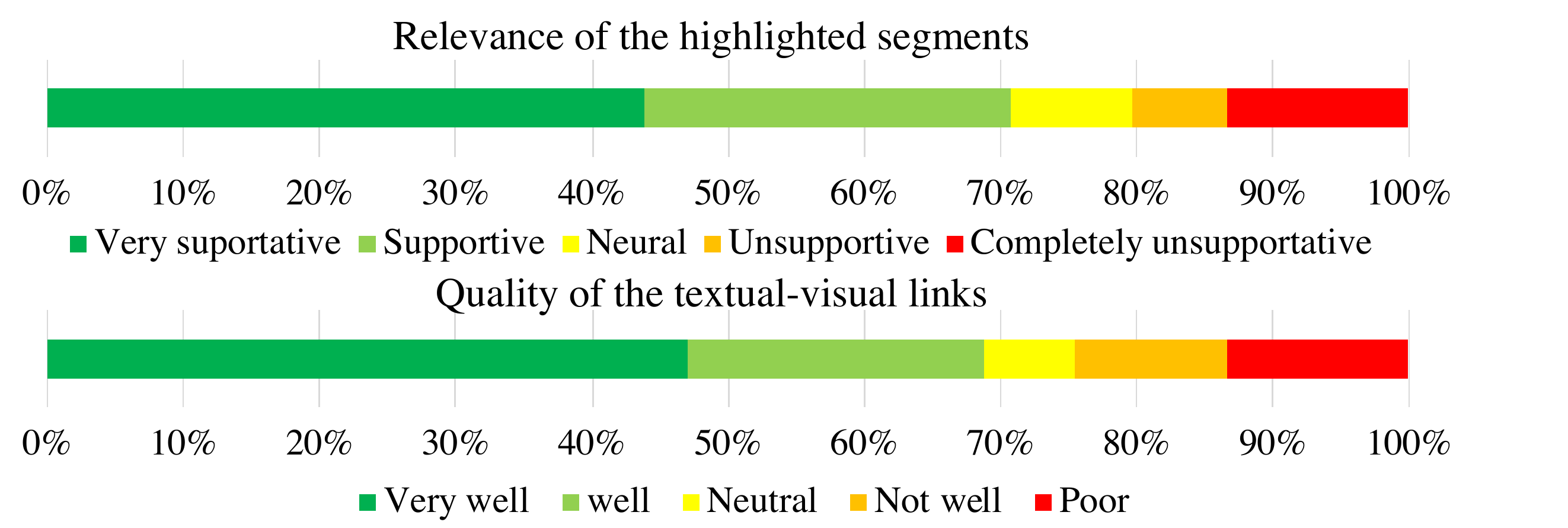}
    \caption{Human evaluation results.  }
    \label{fig:multimodal-eval}
\end{figure}

As shown in Figure \ref{fig:multimodal-eval}, in both cases, about 70\% of the evaluations are positive and about 45\% of them are strongly positive. This indicates that our multimodal explanations provide good connections among visual explanations, textual explanations, and the VQA process. Figure \ref{fig:examples_cheery} presents some sample positively-rated multimodal explanations.

\subsection{Faithfulness Evaluation} 
In this section, we measure the faithfulness of our explanations, i.e.\ how well they reflect the underlying VQA system's reasoning. First, we measured how many words in a generated explanation are actually linked to a visual segmentation in the image. We analyzed the explanations from 1,000 correctly answered questions from the test data. On average, our model is able to link $1.6$ words in an explanation to an image segment, indicating that the textual explanation is actually grounded in objects detected by our VQA system.\\

\noindent\textbf{Faithfulness Evaluation using LIME.} We use the model-agnostic explainer LIME \cite{lime} to determine the segmented objects that most influenced a particular answer, and measure how well the objects referenced in our explanation match these influential segments. We regard all the detected visual segments as the ``interpretable'' units used by LIME to explain decisions.  Using these interpretable units, LIME applies LASSO with the regularization path \cite{efron2004least} to learn a linear model of the local decision boundary around the example to be explained. In particular, we collect $256$ points around the example by randomly blinding each segment's features with a  probability of $0.4$.  The highly weighted features in this model are claimed to provide a faithful explanation of the decision on this example \cite{lime}. The complexity of the explanation is controlled by the number of units, $K$, that can be used in this linear model. 
Using the coefficients $\textbf{w}$ of LIME's weighted linear model, we compare the object segments selected by LIME to the set of objects that are actually linked to words in our explanations.  
Specifically, we define our faithfulness metric as: 
\begin{align}
    score =  \frac{\sum_{i=1}^{|V|}|w_i| \max_{j \in \mathcal{L}}cos(\textbf{v}_i, \textbf{v}_j)}{\sum_{i=1}^{|V|}|w_i|}
    \label{eq:metric1}
\end{align}
where 
$\textbf{v}_i$ denotes the visual feature of the $i$-th segmented object and the $\mathcal{L}$ denotes the set of explanation-linked objects.  For each object in the LIME explanation, it finds the closest object in our explanation and multiplies its LIME weight by this
similarity.   The normalized sum of these matches is used to measure the similarity of the two explanations.  

We collect all correctly answered questions in the VQA-X test set, and Table \ref{tab:faithful_exp0} reports the average score for their explanations using models trained on 15K training explanations with different numbers of interpretable units $K$. The influential objects recognized by LIME match objects that are linked to words in our explanations with an average cosine similarity around $0.7$. This indicates that the explanations are faithfully making reference to visual segmentations that actually influenced the decision of the underlying VQA system. Also, we observe that training with faithful human explanation outperforms purely mimicking human explanations in terms of our faithful metric, and further enforcing the local faithfulness achieves a better result.

\begin{table}[h]
\centering
\begin{tabular}{l|c|c|c}
\toprule
      &K = 1    &  K = 2    &  K = 3  \\ \hline
Ours (Random)  & 0.588 & 0.601 & 0.574 \\
Ours (Filtered) & 0.636 & 0.651 & 0.643 \\
Ours (Filtered+$\mathcal{L}_f$)  & \textbf{0.686} & \textbf{0.705} & \textbf{0.678} \\
\bottomrule
\end{tabular}
\caption{Evaluation of LIME-based faithfulness scores for different numbers of interpretable units $K$ using 15K training explanations. ``Random'' means the training explanations are randomly sampled from the train set, and ``Filtered'' means the models are trained using the remaining faithful explanations.}
\label{tab:faithful_exp0}
\end{table}
\noindent\textbf{Faithfulness Evaluation using Grad-CAM.} We also evaluated the consistency between the Grad-CAM visual explanation vectors from the textual explanation and the predicted answer using the faithful score $\mathcal{S}_f$ defined in Eq.\ \ref{eq:faith_score}.  Table \ref{tab:faithful_gradcam} reports the results from using filtered verses randomly sampled explanations for training. We observe that with faithful human explanations, the average faithfulness evaluation score increases 7\% over training with randomly sampled explanations. Moreover, with the faithfulness loss $\mathcal{L}_f$, the model can better align the visual explanation for the textual explanation with that for the predicted answer, leading to a further 11\% increase.

\begin{figure}[h]
    \centering
    \includegraphics[width=\linewidth, trim={0.5cm 12cm 5.5cm 0cm}, clip]{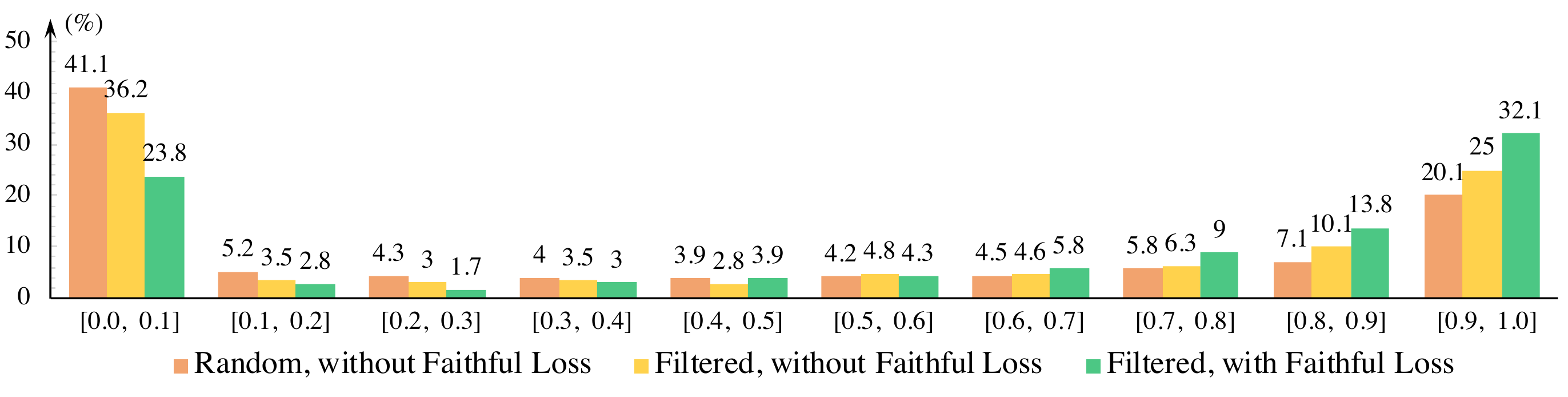}
    \caption{The distribution of explanations' cosine similarity between the visual explanation of the textual explanation and the predicted answer.}
    \label{fig:test_distribution}
\end{figure}

\begin{table}[h]
\centering
\begin{tabular}{l|c|c}
\toprule
     & \# Expl.& Average $\mathcal{S}_f$\\ \hline
Ours (Random) & 15K & 0.38  \\
Ours (Filtered)&15K & 0.45 \\
Ours (Filtered+$\mathcal{L}_f$)&15K & \textbf{0.56}\\
\bottomrule
\end{tabular}
\caption{Average faithfulness evaluation score using different explanations models. ``Random'' means the training explanations are randomly sampled from the train set, and ``Filtered'' means the models are trained using the remaining faithful explanations.}
\label{tab:faithful_gradcam}
\end{table}
We also report the distribution of the generated explanations' cosine similarity between their visual explanation and the visual explanation of the answers in Figure \ref{fig:test_distribution}. The fraction of the faithfulness scores between the interval [0.0,\ 0.1] is significantly decreased by over 17\% when using the faithful human explanations for supervision and further enforcing the local faithfulness with the faithfulness loss, $\mathcal{L}_f$.
\section{Conclusion and Future Work}
This paper has presented a new approach to generating multimodal explanations for visual question answering systems that aims to more faithfully represent the reasoning of the underlying VQA system while maintaining the style of human explanations. The approach generates textual explanations with words linked to relevant image regions actually attended to by the underlying  VQA system.  Experimental evaluations of both the textual and visual aspects of the explanations using both automated metrics and crowdsourced human judgments were presented that demonstrate the advantages of this approach compared to a previously-published competing method. In the future, we would like to incorporate more information from the VQA networks into the explanations. In particular, we would like to integrate {\it network dissection} \cite{netdissect2017} to allow recognizable concepts in the learned hidden-layer representations to be included in explanations. 
\section*{Acknowledgement}
This research was supported by the DARPA XAI program under a grant from AFRL. 

\bibliography{acl2019}
\bibliographystyle{acl_natbib}

\end{document}